\documentclass[journal]{IEEEtran}
\usepackage{amsmath,amssymb}
\usepackage{graphicx}
\usepackage{booktabs}
\usepackage{cite}
\usepackage[hidelinks]{hyperref}
\usepackage{algorithm}
\usepackage{algorithmic}
\usepackage{multirow}
\usepackage{array}
\usepackage{balance}

\begin{document}

\title{Benchmarking Classical Coverage Path Planning Heuristics\\
on Irregular Hexagonal Grids for Maritime Coverage Scenarios}

\author{Carlos~S.~Sep\'ulveda~and~Gonzalo~A.~Ruz, \IEEEmembership{Senior Member, IEEE}%
\thanks{C.~S.~Sep\'ulveda is with the Facultad de Ingenier\'{\i}a y Ciencias,
Universidad Adolfo Ib\'a\~nez, Santiago, Chile, and the Direcci\'on de Programas,
Investigaci\'on y Desarrollo, Armada de Chile, Valpara\'{\i}so, Chile
(carlos.sepulveda@alumnos.uai.cl).}%
\thanks{G.~A.~Ruz is with the Facultad de Ingenier\'{\i}a y Ciencias,
Universidad Adolfo Ib\'a\~nez, Santiago, Chile, the Millennium Nucleus
for Social Data Science (SODAS), and the Millennium Nucleus in Data Science
for Plant Resilience (PhytoLearning) (gonzalo.ruz@uai.cl).}%
\thanks{Supported by the Chilean Navy (Directorate of Programs, Research
and Development), ANID FONDECYT 1230315, ANID-MILENIO-NCN2024\_103,
ANID-MILENIO-NCN2024\_047, CMM FB210005, and ANID Doctorado Nacional 21210465.}}


\maketitle

\begin{abstract}
Coverage path planning on irregular hexagonal grids is relevant to maritime surveillance, search and rescue and environmental monitoring, yet classical methods are often compared on small ad hoc examples or on rectangular grids. This paper presents a reproducible benchmark of deterministic single-vehicle coverage path planning heuristics on irregular hexagonal graphs derived from
synthetic but maritime-motivated areas of interest. The benchmark contains 10\,000 Hamiltonian-feasible instances spanning compact, elongated, and irregular morphologies, 17 heuristics from seven families, and a common evaluation protocol covering Hamiltonian success, complete-coverage success, revisits, path length, heading changes, and CPU latency. Across the released dataset, heuristics with explicit shortest-path reconnection solve the relaxed coverage task reliably but almost never produce zero-revisit tours. Exact Depth-First Search confirms that every released instance is Hamiltonian-feasible. The strongest classical Hamiltonian baseline is a Warnsdorff variant that uses an index-based tie-break together with a terminal-inclusive residual-degree policy, reaching 79.0\% Hamiltonian success. The dominant design choice is not tie-breaking alone, but how the residual degree is defined when the endpoint is reserved until the final move. This shows that underreported implementation details can materially affect performance on sparse geometric graphs with bottlenecks. The benchmark is intended as a controlled testbed for heuristic analysis rather than as a claim of operational optimality at fleet scale.
\end{abstract}


\section{Introduction}\label{sec:intro}

\IEEEPARstart{C}{overage} path planning (CPP) seeks trajectories that observe every point in an area of interest (AOI) while controlling travel effort, redundancy, and maneuvering complexity~\cite{choset2001coverage,Galceran2013, cabreira2019,Fevgas2022}. In ocean and coastal operations, CPP arises in persistent surveillance, search and rescue, environmental monitoring, and
inspection missions performed by unmanned surface, aerial, or underwater vehicles~\cite{Cho2021,Cho2021b}. In these settings, the AOI is rarely convex: coastlines, islands, exclusion zones, and navigational constraints generate narrow passages and bottleneck corridors that challenge simple sweep patterns.

Hexagonal discretizations are attractive for maritime sensing because they provide near-isotropic local connectivity, uniform neighbor distance, and a natural approximation to circular sensing footprints~\cite{birch2007rectangular, Kadioglu2019}. However, most CPP comparisons in the literature either focus on rectangular grids, exact methods on small instances, or case-specific examples that do not support broad conclusions about robustness across irregular topologies~\cite{Galceran2013,cabreira2019,TanChee2021}.

This paper addresses a narrower but practically important question: \emph{how do classical deterministic CPP heuristics compare when they are evaluated under a common protocol on a large set of irregular hexagonal graphs that remain small enough to audit for Hamiltonian feasibility?} We do not propose a new planner. Instead, we contribute a controlled benchmark that makes algorithmic strengths, failure modes, and implementation sensitivities visible.

The main contributions are:
\begin{enumerate}
\item A dataset of 10\,000 irregular hexagonal AOIs with three morphology
families and verified Hamiltonian feasibility.
\item A common implementation and evaluation protocol for 17 deterministic
heuristics from seven families, all operating on the same graph model and start /
return conventions.
\item A comparative analysis covering Hamiltonian success, complete-coverage
success, revisits, path length, heading changes, latency, morphology-specific
behavior, and failure modes.
\item Evidence that \emph{residual-degree policy under endpoint reservation} is a
first-order design choice for Warnsdorff-style traversal on sparse hexagonal
graphs, whereas tie-breaking acts as a secondary effect.
\item Public release of the benchmark generator, audited instances, and
evaluation scripts to support reproducible comparison with future heuristic,
exact, and learning-based methods.
\end{enumerate}

\section{Related Work}\label{sec:related}

Classical CPP surveys organize methods by decomposition and traversal strategy, including cellular decomposition, spanning-tree coverage, and sweep-based patterns~\cite{choset2001coverage,Galceran2013,cabreira2019,Fevgas2022}. Boustrophedon coverage~\cite{choset2001coverage} and spanning-tree coverage~\cite{gabriely2001spanning} remain standard baselines because they provide simple deterministic behavior and strong relaxed-coverage guarantees. Bahnemann et al.~\cite{bahnemann2021revisiting} recently revisited boustrophedon planning as a generalized TSP, highlighting the gap between classical sweep assumptions and practical irregular decompositions.

Hexagonal decompositions have been explored in robotics and UAV coverage because their geometry reduces directional bias and supports natural neighborhood relations~\cite{birch2007rectangular,Azpurua2018,Kadioglu2019}. In maritime settings, hexagonal grids also appear in exact or mixed-integer formulations for multi-UAV search and rescue~\cite{Cho2021,Cho2021b}. Those works are important for operational motivation, but they do not provide a large-scale benchmark of classical heuristics on irregular sparse hexagonal graphs.

Warnsdorff's rule, originally proposed for the Knight's Tour~\cite{Warnsdorff1823} and later generalized to Hamilton paths on graphs~\cite{pohl1967method}, is a natural degree-based heuristic for sparse traversal problems. In the specific context of hexagonal grid graphs, Hamiltonicity has also been studied from a graph-theoretic perspective~\cite{islam2007hamilton}. The appeal of Warnsdorff-style traversal lies in the simplicity of the residual-degree heuristic. Its weakness is that local choices can irreversibly consume articulation-like corridor cells in irregular topologies. The present benchmark shows that Warnsdorff is not fully specified by ``choose the minimum residual degree'' alone when the terminal node is reserved until the final move: the operational definition of residual degree itself matters.

Recent combinatorial-optimization benchmarks emphasize Euclidean routing on complete graphs~\cite{berto2025rl4co}. CPP on sparse geometric graphs is structurally different: adjacency constraints, obstacle-induced bottlenecks, and the distinction between relaxed coverage and zero-revisit traversal are central. A dedicated benchmark is therefore justified.

\section{Problem Formulation}\label{sec:problem}

\subsection{Hexagonal AOI Graph}

Let $\mathcal{A}\subset\mathbb{R}^2$ denote a polygonal area of interest (AOI), possibly with interior holes representing forbidden or obstructed regions. Rather than using an exact continuous-space decomposition, we adopt a fixed-resolution approximate cellular decomposition based on a regular hexagonal lattice. The lattice is aligned with the minimum rotated rectangle of $\mathcal{A}$ and instantiated as an even-$q$ offset grid with hexagon circumradius $h$.

A hexagonal cell is retained if it has sufficient geometric support in the free space according to intersection and overlap tests with the AOI and its holes. The resulting binary occupancy mask is then post-processed to remove discretization artifacts: we keep the largest connected component, remove interior dead ends, preserve a single exterior boundary ring, and enforce accessibility from the exterior border. This post-processing suppresses discretization-induced topological
artifacts so that benchmark failures are attributable to heuristic behavior rather than to spurious graph defects. This yields a set of visitable cells $\mathcal{V}$ with $|\mathcal{V}|\in[28,46]$.

An undirected graph $G=(\mathcal{V}\cup\{b,b'\},\mathcal{E})$ is finally constructed by connecting face-adjacent visitable hexagons. A departure node $b$ and a return node $b'$ are attached only to cells on the outer ring that admit a feasible line-of-sight connection from the launch location. We emphasize that this representation is intended to define a controlled family of sparse hexagonal graphs for reproducible benchmarking and exact Hamiltonian auditing, rather than to claim an exact geometric decomposition of the continuous AOI.

The choice of a regular hexagonal discretization is deliberate. Relative to square grids, hexagonal cells provide a more isotropic local neighborhood and a natural surrogate for approximately circular sensing footprints, while still yielding graphs of manageable size for exhaustive feasibility auditing.
Conversely, exact cellular decompositions are often preferable when continuous boundary fidelity is the primary concern. In this work, our objective is not to optimize the decomposition itself, but to compare CPP heuristics under a common, reproducible graph representation.

\subsection{Two Coverage Objectives}

We distinguish two tasks.

\textit{Relaxed coverage} requires a walk from $b$ to $b'$ that visits every cell in $\mathcal{V}$ at least once; revisits are allowed.

\textit{Hamiltonian coverage} requires a path $\pi=(b,v_1,\ldots,v_{|\mathcal{V}|},b')$ that visits each cell exactly once. This is the zero-revisit version of the problem and is combinatorially much more difficult on sparse irregular graphs~\cite{islam2007hamilton}.

\subsection{Benchmark Perspective}

The benchmark is intentionally focused on graphs small enough to permit exact Hamiltonian feasibility auditing, yet rich enough to contain narrow corridors, concavities, and obstacle-induced dead ends. The goal is not to claim that these instances represent the full scale of operational missions, but to provide a controlled regime in which qualitative differences among heuristics can be measured reproducibly.

\section{Dataset}\label{sec:dataset}

\subsection{Generation Pipeline}

Each instance is generated in three stages: (i) an outer polygon is sampled from one of three morphology families; (ii) a hexagonal tessellation is mounted in the oriented bounding-box frame; and (iii) interior cells are removed to create islands, shoals, exclusion zones, and bottleneck corridors.

Morphology is classified post hoc using the Polsby--Popper compactness ratio $c = 4\pi A/P^2$ and the aspect ratio $\alpha$ of the minimum rotated rectangle of the AOI polygon:
\textbf{Compact} instances ($c > 0.6$ and $\alpha < 2$) are convex or near-convex shapes representing open patrol zones.
\textbf{Elongated} instances ($\alpha \geq 2$) are high-aspect-ratio shapes
representing coastal strips, channels, or fjord-like corridors.
\textbf{Irregular} instances (all remaining, i.e., $c \leq 0.6$ and $\alpha < 2$) are concave shapes with deeper indentations and stronger obstacle interaction.
The released dataset contains 5\,788 compact, 177 elongated, and 4\,035 irregular instances.

\begin{figure*}[t]
\centering
\begin{minipage}{0.18\textwidth}
    \centering
    \includegraphics[width=\linewidth]{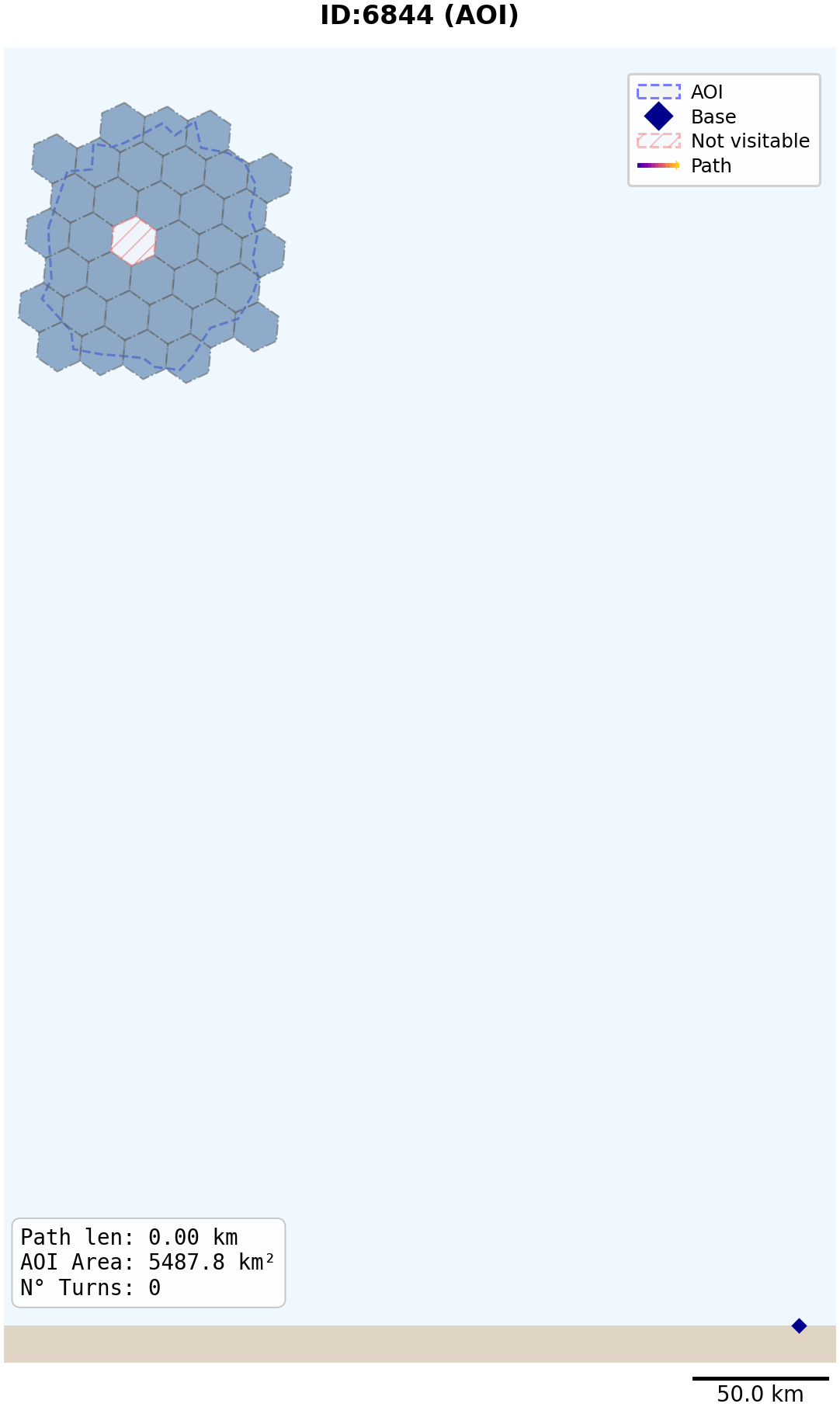}
    \small Compact ($c{=}0.78$, $\alpha{=}1.3$)
\end{minipage}
\hfill
\begin{minipage}{0.48\textwidth}
    \centering
    \includegraphics[width=\linewidth]{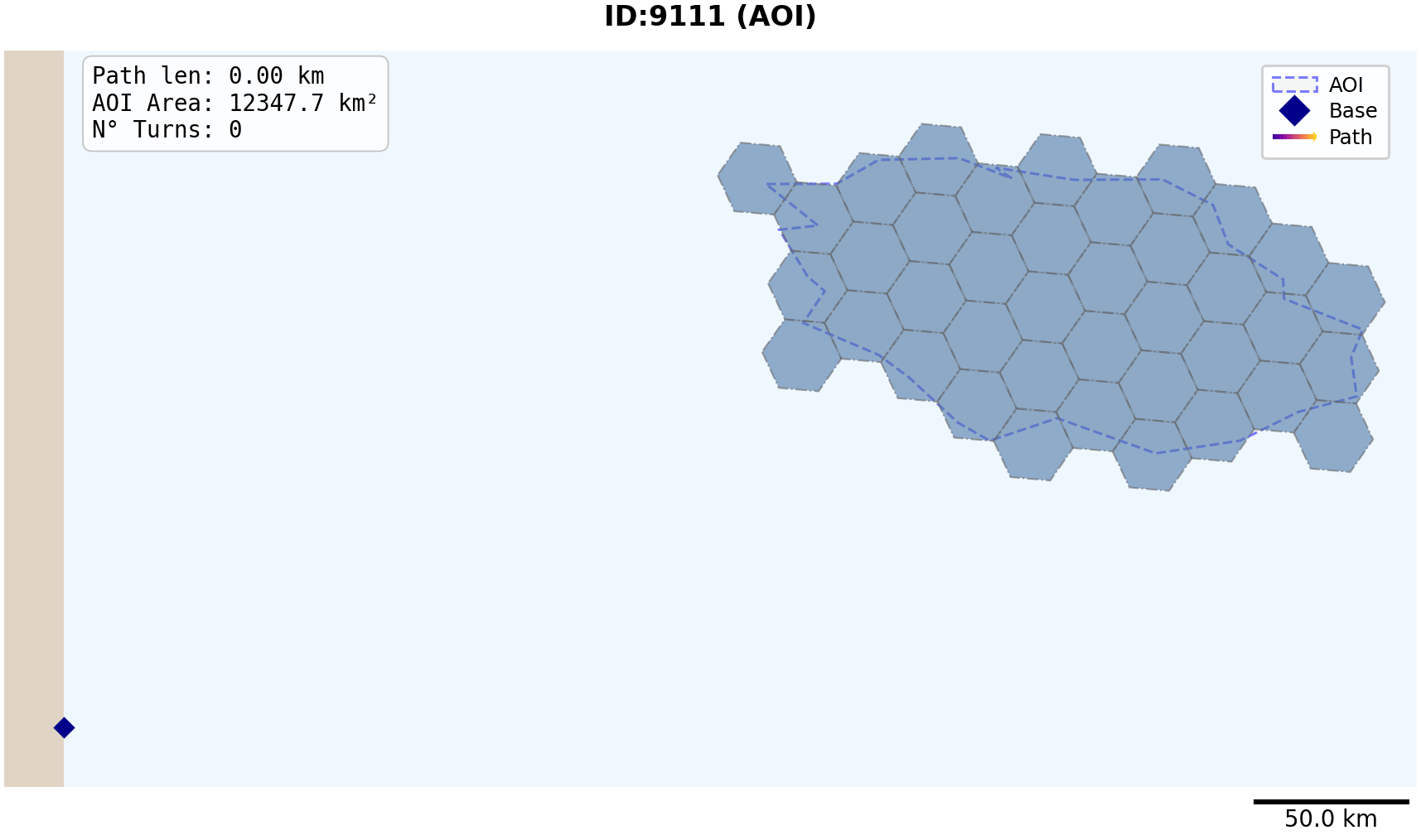}
    \small Elongated ($\alpha{=}3.1$)
\end{minipage}
\hfill
\begin{minipage}{0.30\textwidth}
    \centering
    \includegraphics[width=\linewidth]{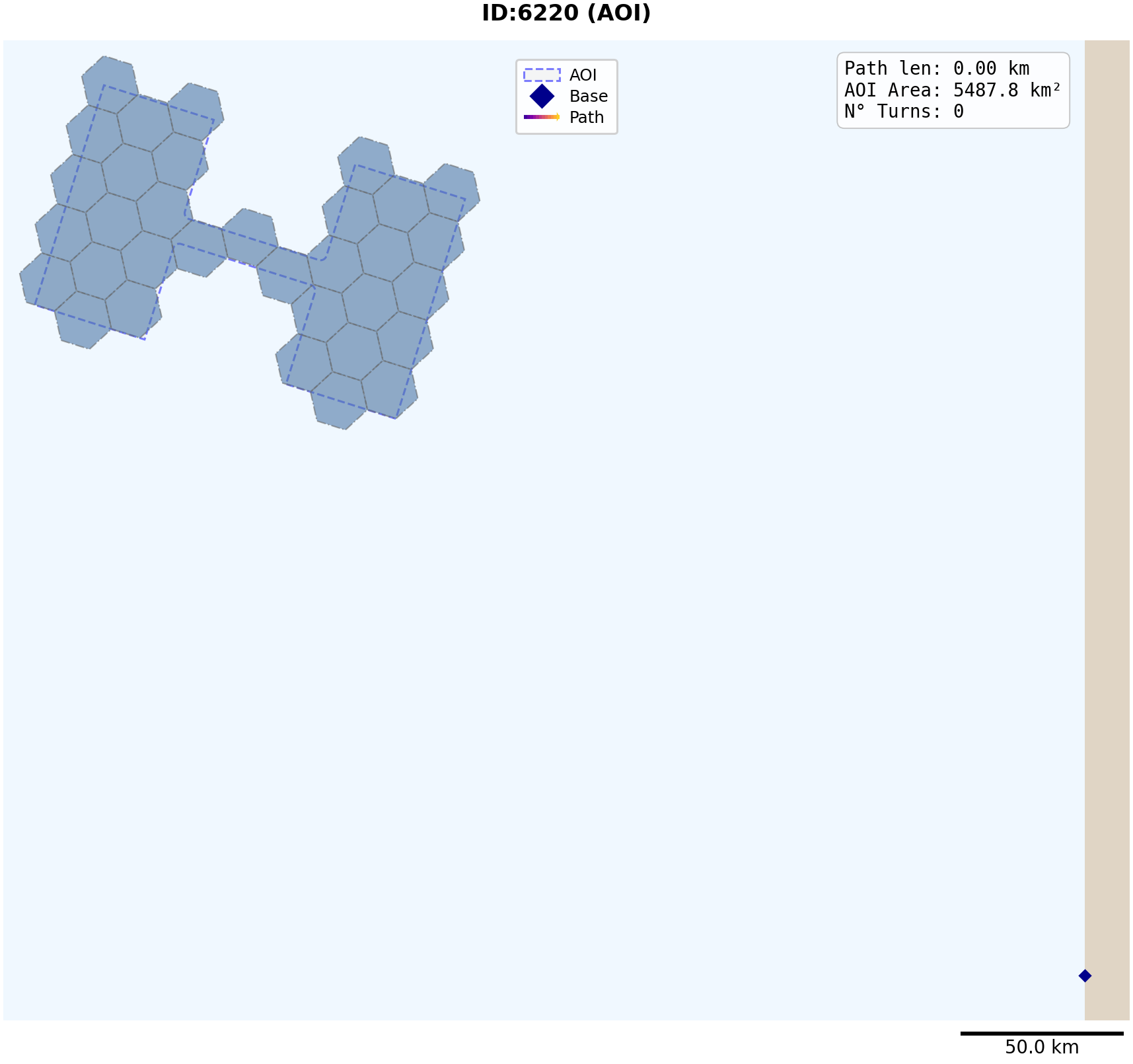}
    \small Irregular with narrow corridor ($c{=}0.41$, $\alpha{=}1.6$)
\end{minipage}

\vspace{2mm}
\caption{Representative instances from each morphological family showing the hexagonal tessellation with base node ($\diamondsuit$) serving as both starting and terminal point.}
\label{fig:morphology}
\end{figure*}

\subsection{Maritime Motivation and Scale}

The geometric scales are maritime-motivated rather than claimed to be fully representative of any single platform or theater. AOI areas, hex-cell sizes, and base standoff distances were selected to create sparse graphs in the 28--46-cell range while preserving an interpretation compatible with maritime coverage scenarios. This choice keeps the benchmark computationally auditable and
exposes the tension between complete coverage and zero-revisit traversal.

\subsection{Feasibility Audit}

Every released instance was audited offline for Hamiltonian feasibility using exhaustive Depth-First Search (DFS) with backtracking. Instances failing the audit were discarded. This design choice is essential: a heuristic should fail on this benchmark because of algorithmic weakness, not because the underlying instance is inherently infeasible.

\subsection{Evaluation Protocol}

All 17 heuristics are deterministic and are evaluated on the full 10\,000-instance dataset. Every method receives the same graph, the same start and return convention, and the same adjacency constraints. No method is given instance-specific tuning. Because the benchmark evaluates the full released dataset rather than a random sample, we focus on descriptive statistics over the entire benchmark.

\section{Heuristic Algorithms}\label{sec:heuristics}

All heuristics operate on the same graph $G$ and are grouped into seven families. Full implementations are provided in the released repository; we describe each family below and provide pseudocode for the Warnsdorff variants (Algorithm~\ref{alg:warnsdorff}), which are central to the paper's main finding.

\subsection{Family 1: Linear Sweep (3 methods)}

\textbf{Boustrophedon} projects centroids onto the principal axis of the oriented bounding box, partitions them into rows, and traverses consecutive rows in alternating directions. Inter-row transitions use shortest-path reconnections on the adjacency graph, which may revisit cells.
\textbf{Row-OneWay} retains a fixed row direction and uses fly-back transitions. \textbf{Segment-Snake} applies the same idea within obstacle-aware local segments.

\subsection{Family 2: Interleaved Sweep (2 methods)}

\textbf{Row-Interleave} visits rows in an interleaved order to reduce sharp local turn sequences. \textbf{Seg.-Interleave} applies the same principle within obstacle-aware segments.

\subsection{Family 3: Contour / Spiral (3 methods)}

\textbf{Spiral-Inward} traverses successive boundary rings from the perimeter inward. \textbf{Spiral-Outward} starts near the center and expands outward.
\textbf{Boundary-Peel} repeatedly removes the current outer layer and reconnects when necessary.

\subsection{Family 4: Spanning-Tree Coverage (2 methods)}

\textbf{STC-Tree} follows the Spanning Tree Coverage paradigm~\cite{gabriely2001spanning}: a spanning tree of the internal-cell
graph is constructed and then circumnavigated via DFS to guarantee complete coverage. Gabriely and Rimon showed that any spanning tree suffices for the coverage guarantee; the choice of tree affects path quality but not completeness. In our hexagonal adaptation, the spanning tree is built by BFS rooted at the internal cell nearest to the base; if the BFS tree does not span all internal cells, the instance is reported as unsolved. \textbf{STC-Like} uses the same DFS circumnavigation but builds the spanning tree with a distance-biased BFS that expands nearest neighbors first, producing shorter tree edges. Additionally, any internal cells not reached by the initial tree are
connected via greedy nearest-neighbor shortest-path bridges after the circumnavigation, making this variant more robust to weakly connected subgraphs.

\subsection{Family 5: Graph-Based Local Search (5 methods)}\label{sec:warn_detail}

\subsubsection{Warnsdorff Variants}

Warnsdorff's rule selects the unvisited neighbor with minimum residual degree, excluding the return node until all internal cells have been visited. The benchmark distinguishes two independent implementation axes.

First, two tie-break rules are evaluated: \textbf{index} (prefer the smallest node index) and \textbf{distance} (prefer the smallest Euclidean move from the current node).

Second, two residual-degree policies are evaluated under endpoint reservation.
In the \textbf{endpoint-aware (EP)} policy, the terminal node is excluded from the residual-degree count while more than one internal target remains; the resulting degree therefore reflects only future moves that remain effectively usable during the current phase of the traversal. In the
\textbf{terminal-inclusive (TI)} policy, the terminal remains excluded from the candidate set but is still counted inside the residual degree, so endpoint adjacency acts as a latent structural signal rather than as an immediately admissible move. Combining these two axes yields four Warnsdorff variants:
Warnsdorff-EP (index), Warnsdorff-EP (dist.), Warnsdorff-TI (index), and
Warnsdorff-TI (dist.).

Algorithm~\ref{alg:warnsdorff} shows the unified procedure. The critical
distinction between EP and TI is in line~8: under EP, the terminal $b'$ is
excluded from the neighbor count of candidate $j$ when unvisited targets
remain, so the degree reflects only immediately usable moves. Under TI, $b'$ is
counted in the degree even though it is not an admissible candidate, providing a
latent proximity signal that discourages premature consumption of cells adjacent
to the terminal.

Because Warnsdorff variants have no backtracking mechanism, a failure to find
an unvisited neighbor at any step causes immediate termination. Consequently,
for all four Warnsdorff variants the coverage completion rate (CCR) equals the
Hamiltonian success rate (HSR).

\begin{algorithm}[t]
\small
\caption{Warnsdorff's Rule (unified EP/TI formulation)}
\label{alg:warnsdorff}
\begin{algorithmic}[1]

\REQUIRE Graph $G$, base $b$, terminal $b'$, 
tie\_break $\in\{\text{index},\text{distance}\}$,
policy $\in\{\text{EP},\text{TI}\}$

\STATE $v \gets b$;\; visited $\gets \{b\}$;\; $\pi \gets [b]$

\WHILE{unvisited internal targets remain}

  \STATE $\mathcal{C} \gets \{j\in\mathcal{N}(v) : j\notin\text{visited} \text{ and } (j\neq b' \text{ if targets remain})\}$

  \IF{$\mathcal{C}=\emptyset$}
    \RETURN $\text{(}\pi,\text{ FAIL)}$
  \ENDIF

  \FOR{each $j\in\mathcal{C}$}

    \STATE $d_j \gets |\{k\in\mathcal{N}(j) : k\notin(\text{visited}\cup\{v\}) \text{ and } (k\neq b' \text{ if policy=EP and targets $> 1$})\}|$

    \IF{tie\_break = index}
      \STATE $t_j \gets j$
    \ELSE
      \STATE $t_j \gets \|\mathbf{s}_v - \mathbf{s}_j\|_2$
    \ENDIF
  \ENDFOR
  \STATE $v^* \gets \arg\min_{j\in\mathcal{C}} (d_j,\,t_j)$
  \STATE visited $\gets$ visited $\cup\{v^*\}$;\;
    $\pi \gets \pi\oplus v^*$;\; $v \gets v^*$
\ENDWHILE
\IF{$b'\in\mathcal{N}(v)$}
  \RETURN $\text{(}\pi\oplus b',\text{ SUCCESS)}$
\ELSE
  \RETURN $\text{(}\pi,\text{ FAIL)}$
\ENDIF

\end{algorithmic}
\end{algorithm}

\subsubsection{DFS-Backtrack}

DFS-Backtrack greedily extends the path through unvisited neighbors, selecting the neighbor with minimum residual degree at each step. When none is available, it backtracks on the visited subgraph to the nearest node that still has an unexplored branch via BFS shortest path, incurring revisits. After covering all internal cells it returns to the terminal via shortest path.

\subsection{Family 6: Wavefront / Brushfire (1 method)}

\textbf{Wavefront-Hex} constructs a static distance field on the internal hex
graph via multi-source BFS seeded at the terminal frontier (internal cells
adjacent to $b'$). Each internal cell receives a wavefront label equal to its
BFS distance from this frontier. The traversal proceeds greedily from $b$: at
each step, it moves to the adjacent unvisited cell that maximizes the wavefront
label (farthest from the terminal), with ties broken lexicographically by
minimum residual unvisited degree, minimum Euclidean distance, and minimum node
index. When no adjacent unvisited cell exists, a BFS connector through the
internal subgraph (traversing both visited and unvisited cells) bridges to the
highest-wavefront remaining cell reachable via the shortest path. Cells
traversed by the connector are marked as covered. This mechanism guarantees
complete coverage but introduces revisits when the connector passes through
previously visited cells.

\subsection{Family 7: Space-Filling Curve (1 method)}

\textbf{Morton Z-order} ranks centroids by Morton code (bit-interleaved
quantized coordinates) and reconnects non-adjacent successive cells through
shortest feasible paths.

\section{Evaluation Metrics}\label{sec:metrics}

We report the following metrics.

\textbf{Hamiltonian success rate (HSR)}: fraction of instances solved with a
valid zero-revisit path from $b$ to $b'$.

\textbf{Coverage completion rate (CCR)}: fraction of instances in which all
cells are covered at least once.

\textbf{Revisits}: mean and standard deviation of repeated cell visits on the
subset of completed-coverage instances.

\textbf{Distance}: mean and standard deviation of the total Euclidean path length in the normalized coordinate system of the graph (coordinates are base-centered and scaled by the maximum cell-to-base radius per instance), computed on the subset of completed-coverage instances.

\textbf{Turns}: mean and standard deviation of the cumulative absolute heading change (in radians) along the path, computed on the subset of completed-coverage instances.

\textbf{Latency}: CPU wall-clock time in milliseconds per instance, measured on
an AMD Ryzen 9 5900HX.

\section{Results}\label{sec:results}

All results are computed on the full 10\,000-instance benchmark.

\subsection{Feasibility}

\begin{table}[t]
\centering
\caption{Feasibility on the full benchmark. HSR: Hamiltonian success rate. CCR:
complete-coverage success rate.}
\label{tab:coverage}
\small
\begin{tabular}{lccc}
\toprule
Method & Family & HSR (\%) & CCR (\%) \\
\midrule
Exact DFS (oracle) & Oracle & 100.0 & 100.0 \\
\midrule
Warnsdorff-TI (index) & Graph & 79.0 & 79.0 \\
Warnsdorff-TI (dist.) & Graph & 71.8 & 71.8 \\
Warnsdorff-EP (index) & Graph & 47.5 & 47.5 \\
DFS-Backtrack & Graph & 34.7 & 100.0 \\
Warnsdorff-EP (dist.) & Graph & 31.0 & 31.0 \\
Boustrophedon & Linear Sweep & 0.0 & 100.0 \\
Row-OneWay & Linear Sweep & 0.0 & 100.0 \\
Segment-Snake & Linear Sweep & 0.0 & 100.0 \\
Row-Interleave & Interleaved & 0.0 & 100.0 \\
Seg.-Interleave & Interleaved & 0.0 & 100.0 \\
Spiral-Outward & Contour & 0.0 & 100.0 \\
Spiral-Inward & Contour & 0.0 & 100.0 \\
Boundary-Peel & Contour & 0.0 & 100.0 \\
STC-Tree & STC & 0.0 & 100.0 \\
STC-Like & STC & 0.0 & 100.0 \\
Wavefront-Hex & Wavefront & 7.3 & 100.0 \\
Morton Z-order & Space-filling & 0.0 & 100.0 \\
\bottomrule
\end{tabular}
\end{table}

Table~\ref{tab:coverage} reveals two robust patterns. First, heuristics that
allow shortest-path reconnection solve the relaxed coverage task reliably but
almost never induce zero-revisit tours. Second, the only nonzero Hamiltonian
success rates are achieved by the Warnsdorff family, DFS-Backtrack, and, to a
much smaller extent, Wavefront-Hex. The strongest classical Hamiltonian
baseline is Warnsdorff-TI (index) at 79.0\% HSR. Note that the four Warnsdorff
variants report HSR~=~CCR because these heuristics lack a backtracking mechanism
and terminate immediately upon encountering an empty candidate set
(Sec.~\ref{sec:warn_detail}).

\subsection{Path Quality}

\begin{table}[t]
\centering
\caption{Path-quality and latency metrics. Revisits, distance, and turns are
computed on the subset of completed-coverage instances for each method.}
\label{tab:quality}
\scriptsize
\resizebox{\columnwidth}{!}{
\begin{tabular}{lcccc}
\toprule
Method & Revisits & Distance (norm.) & Turns (rad) & Latency (ms) \\
\midrule
Warnsdorff-EP (dist.) & $0.0{\pm}0.0$ & $3.23{\pm}0.36$ & $45.1{\pm}5.8$ & 0.62 \\
Warnsdorff-TI (dist.) & $0.0{\pm}0.0$ & $3.26{\pm}0.37$ & $47.1{\pm}6.2$ & 0.64 \\
Warnsdorff-TI (index) & $0.0{\pm}0.0$ & $3.27{\pm}0.38$ & $42.5{\pm}6.5$ & 0.65 \\
Warnsdorff-EP (index) & $0.0{\pm}0.0$ & $3.32{\pm}0.38$ & $42.3{\pm}5.7$ & 0.75 \\
DFS-Backtrack & $3.3{\pm}6.2$ & $3.33{\pm}0.45$ & $49.6{\pm}12.5$ & 1.04 \\
Boustrophedon & $6.4{\pm}2.0$ & $3.56{\pm}0.43$ & $51.6{\pm}6.6$ & 27.39 \\
Segment-Snake & $6.5{\pm}2.1$ & $3.56{\pm}0.43$ & $51.7{\pm}6.5$ & 26.89 \\
Row-OneWay & $8.6{\pm}3.6$ & $3.67{\pm}0.49$ & $54.6{\pm}7.1$ & 28.57 \\
Seg.-Interleave & $16.2{\pm}4.2$ & $3.98{\pm}0.57$ & $61.2{\pm}8.0$ & 30.99 \\
Row-Interleave & $18.4{\pm}4.3$ & $4.04{\pm}0.55$ & $54.0{\pm}7.3$ & 30.00 \\
Spiral-Inward & $12.1{\pm}12.7$ & $3.84{\pm}0.74$ & $50.9{\pm}19.1$ & 1.77 \\
Spiral-Outward & $12.0{\pm}12.6$ & $3.85{\pm}0.74$ & $50.7{\pm}18.2$ & 1.95 \\
Boundary-Peel & $16.1{\pm}12.6$ & $3.85{\pm}0.71$ & $59.9{\pm}17.8$ & 1.29 \\
STC-Tree & $35.8{\pm}3.5$ & $4.61{\pm}0.72$ & $75.6{\pm}13.0$ & 1.04 \\
STC-Like & $35.8{\pm}3.5$ & $4.61{\pm}0.72$ & $78.9{\pm}11.9$ & 0.91 \\
Wavefront-Hex & $5.5{\pm}3.5$ & $3.40{\pm}0.43$ & $40.2{\pm}7.3$ & 2.15 \\
Morton Z-order & $18.6{\pm}5.4$ & $4.13{\pm}0.61$ & $72.7{\pm}10.0$ & 1.36 \\
\bottomrule
\end{tabular}}
\end{table}

Table~\ref{tab:quality} shows that the most attractive heuristic depends on the
objective. Linear sweeps are the strongest baselines when the priority is short
and low-redundancy relaxed coverage. Tree-based coverage methods incur the
largest redundancy because branch circumnavigation is built into their design.
DFS-Backtrack occupies an intermediate position: it can occasionally realize
Hamiltonian traversals, but its relaxed-coverage paths are less structured than
those of the best sweep-based methods. Wavefront-Hex is notable because it
retains a modest Hamiltonian success rate (7.3\%) while also keeping revisits,
distance, and turns relatively low among the revisit-allowed planners.

\subsection{Warnsdorff Sensitivity to Residual-Degree Policy and Tie-Breaking}\label{sec:warn}

\begin{figure*}[htbp]
\centering

\begin{minipage}{0.48\linewidth}
    \centering
    \includegraphics[width=\linewidth]{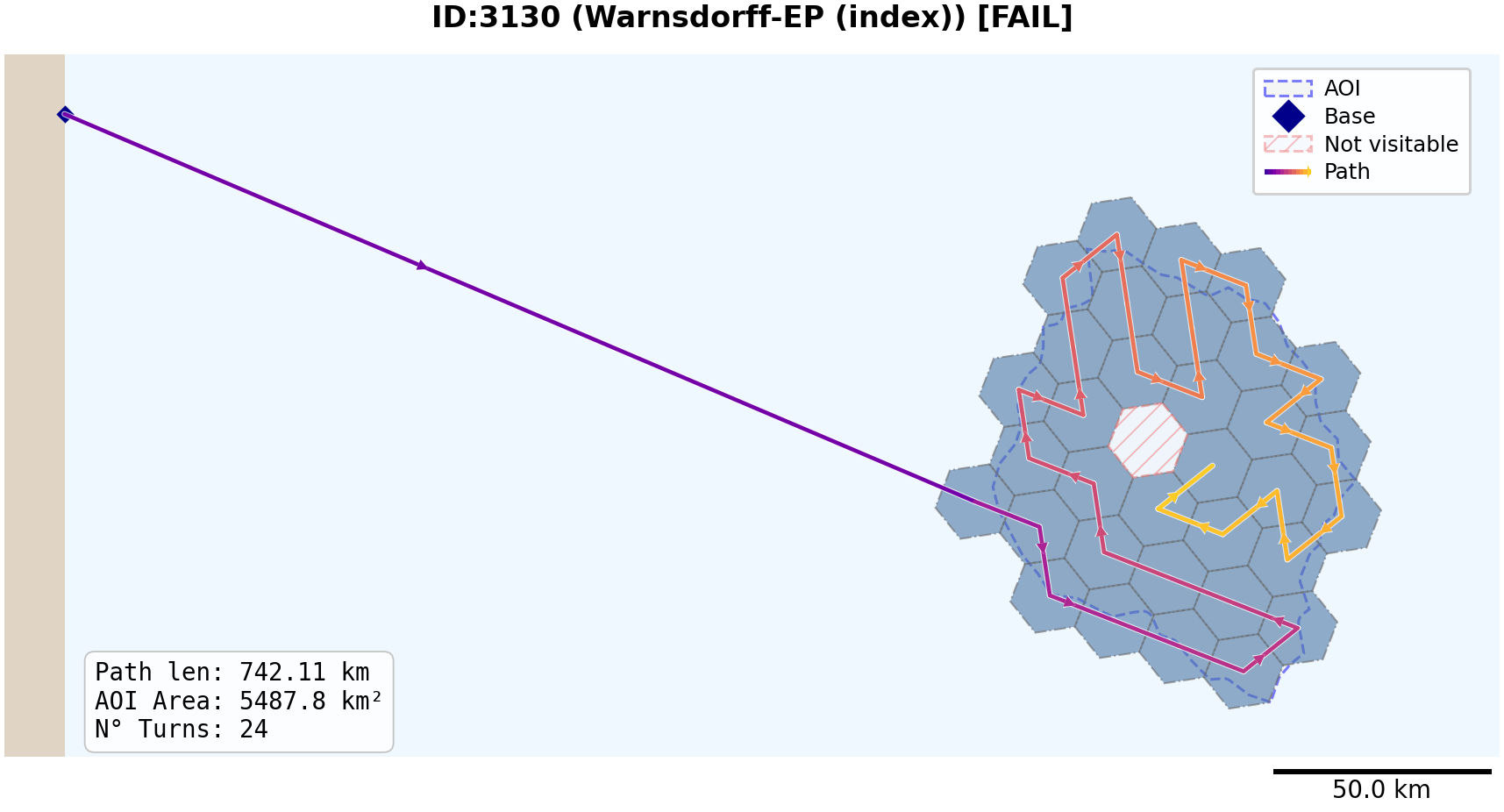}
    \caption*{Warnsdorff-EP (FAIL)}
\end{minipage}
\hfill
\begin{minipage}{0.48\linewidth}
    \centering
    \includegraphics[width=\linewidth]{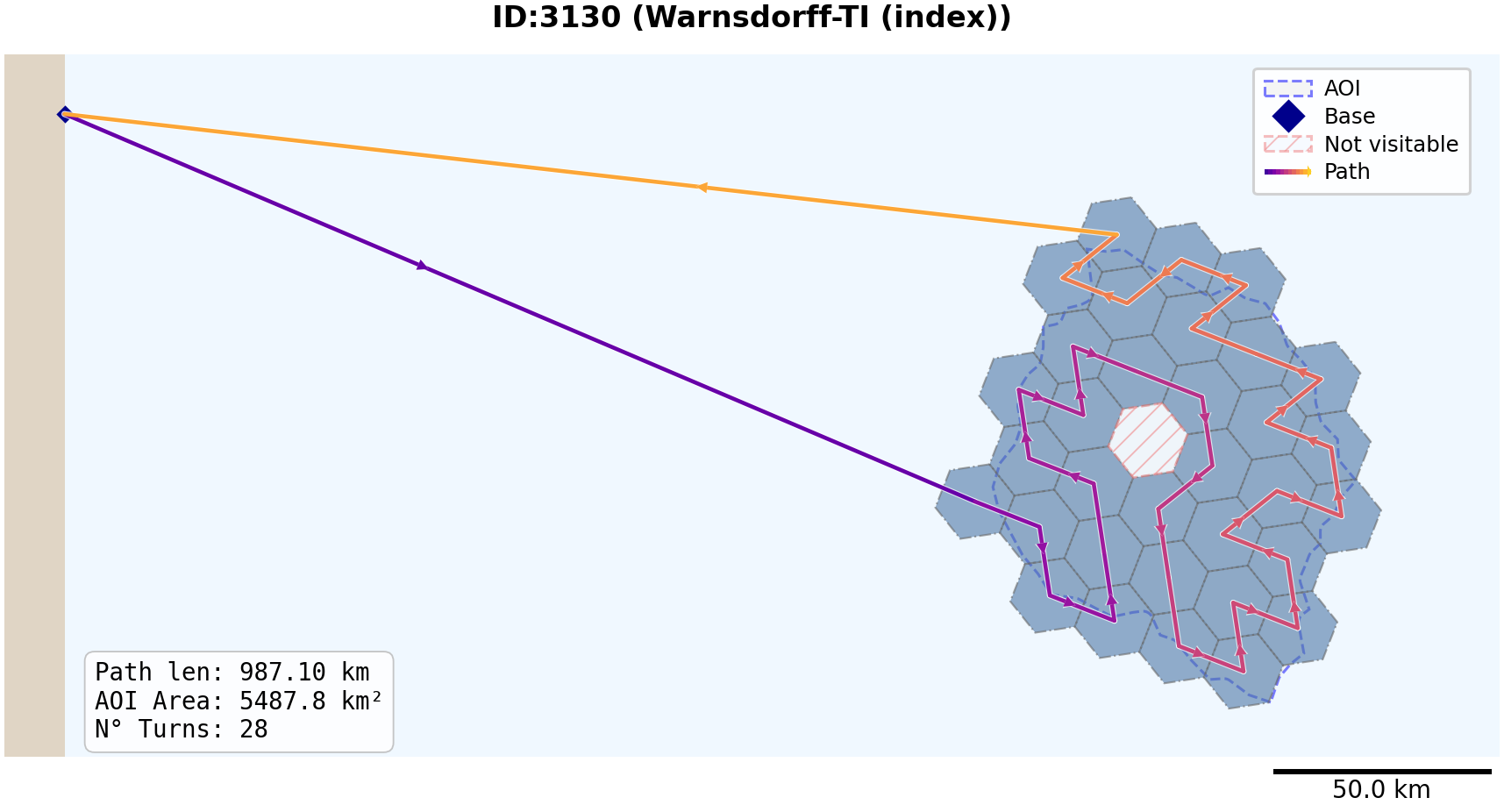}
    \caption*{Warnsdorff-TI (SUCCESS)}
\end{minipage}

\caption{Corridor-consumption failure under EP and its mitigation under TI.
Both use index tie-breaking on the same irregular instance.}
\label{fig:ep_vs_ti}
\end{figure*}

\begin{table}[t]
\centering
\caption{Warnsdorff variants. EP: endpoint-aware residual degree. TI:
terminal-inclusive residual degree.}
\label{tab:warnsdorff}
\small
\resizebox{\linewidth}{!}{
\begin{tabular}{lcccc}
\toprule
Variant & HSR (\%) & Distance (norm.) & Turns (rad) & Latency (ms) \\
\midrule
Warnsdorff-EP (index) & 47.5 & $3.32{\pm}0.38$ & $42.3{\pm}5.7$ & 0.75 \\
Warnsdorff-EP (dist.) & 31.0 & $3.23{\pm}0.36$ & $45.1{\pm}5.8$ & 0.62 \\
Warnsdorff-TI (index) & 79.0 & $3.27{\pm}0.38$ & $42.5{\pm}6.5$ & 0.65 \\
Warnsdorff-TI (dist.) & 71.8 & $3.26{\pm}0.37$ & $47.1{\pm}6.2$ & 0.64 \\
\bottomrule
\end{tabular}}
\end{table}

Table~\ref{tab:warnsdorff} shows that the dominant effect is the residual-degree
policy, not tie-breaking alone. Switching from endpoint-aware to
terminal-inclusive residual degree increases HSR by 31.5 percentage points for
the index tie-break and by 40.8 points for the distance tie-break. Within both
policies, however, index-based tie-breaking outperforms distance-based
tie-breaking on this dataset: by $+$16.5~pp under EP and $+$7.2~pp under TI.

The superiority of index-based tie-breaking is initially counterintuitive,
since Euclidean proximity might be expected to improve locality. A plausible
interpretation is that distance-based tie-breaking makes the traversal overly
myopic: by repeatedly preferring the nearest admissible cell, it may consume
corridor cells before the opposite subregion has been secured. In contrast,
index-based tie-breaking induces a deterministic ordering inherited from the
lattice construction rather than an explicit nearest-neighbor bias. On this
dataset, that ordering appears to be less damaging to connector preservation.

More importantly, the dominant effect is the residual-degree policy under
endpoint reservation. Whether terminal adjacency contributes to the
residual-degree score substantially changes the ranking of candidate moves and,
therefore, the likelihood of preserving narrow connector cells until the final
stages of the traversal.

\subsection{Morphology-Specific Performance}

\begin{table}[t]
\centering
\caption{Hamiltonian success rate of the four Warnsdorff variants stratified by
AOI morphology.}
\label{tab:morphology}
\small
\resizebox{\columnwidth}{!}{
\begin{tabular}{lrcccc}
\toprule
Morphology & $n$ & EP (idx) & EP (dist.) & TI (idx) & TI (dist.) \\
\midrule
Compact & 5788 & 61.1\% & 32.1\% & 91.1\% & 84.0\% \\
Elongated & 177 & 69.5\% & 30.5\% & 77.4\% & 63.3\% \\
Irregular & 4035 & 27.0\% & 29.4\% & 61.7\% & 54.8\% \\
\midrule
Overall & 10000 & 47.5\% & 31.0\% & 79.0\% & 71.8\% \\
\bottomrule
\end{tabular}}
\end{table}

Table~\ref{tab:morphology} underscores why aggregate performance alone is
insufficient. Compact instances are easiest for all Warnsdorff variants, but
the ranking of the four variants is stable across morphology families: the
terminal-inclusive policy dominates the endpoint-aware policy, and index-based
tie-breaking dominates distance-based tie-breaking within each policy. The gap
widens on irregular instances, where articulation-like corridor cells are more
common and local degree information becomes more fragile. The elongated category
($n=177$, 1.8\% of the dataset) is included for completeness; its smaller
sample size means that its individual statistics carry higher variance and
should be interpreted with appropriate caution.

\subsection{Failure-Mode Analysis}

Manual inspection of failures shows a recurring mechanism: the heuristic enters a subregion through a narrow corridor, exhausts that region, and then discovers that the corridor needed to reach the remaining cells has already been consumed.
This failure is especially damaging for greedy Hamiltonian heuristics because no revisit is permitted to recover connectivity. Endpoint-aware Warnsdorff is more vulnerable to this mechanism because it suppresses the endpoint signal inside the residual-degree score. The terminal-inclusive policy partially mitigates the problem by allowing endpoint adjacency to influence candidate ranking earlier, without actually permitting premature selection of the terminal node.

\section{Discussion}\label{sec:discussion}

The benchmark highlights a central trade-off relevant to maritime planning:
methods that are excellent at guaranteed relaxed coverage are not necessarily
useful when zero-revisit behavior matters, and methods with some Hamiltonian
potential can be fragile on obstacle-rich sparse graphs. The right baseline
therefore depends on the operational objective. If the mission tolerates
revisits, structured sweeps remain competitive because of their simplicity and
path economy. If zero-revisit traversal matters, degree-aware graph heuristics
deserve attention, but their implementation details must be reported explicitly.

The Warnsdorff results are the clearest methodological lesson of the benchmark. On sparse irregular graphs with a reserved endpoint, ``minimum residual degree'' is not a fully specified heuristic until the treatment of the terminal node is made explicit. The residual-degree policy has a much larger effect than the secondary tie-break rule on this dataset. This matters for reproducibility:
papers that describe only the headline heuristic but not the endpoint-handling policy risk reporting results that cannot be faithfully replicated.

\textbf{Limitations.} The benchmark is single-agent, static, and synthetic. It does not model currents, kinematic feasibility beyond geometric adjacency, uncertain detection, or online replanning. The graph sizes are deliberately limited to 28--46 cells to preserve exact feasibility auditing; larger instances may change the absolute ranking of methods. These limitations do not invalidate the benchmark, but they define its scope: controlled comparison of classical heuristics on audited irregular hexagonal graphs. The reported results are conditioned on the adopted hexagonal discretization and its post-processing pipeline. Different discretizations (e.g., exact
cellular decomposition, adaptive quadtrees, or triangulations) could alter the absolute difficulty and ranking of the benchmark instances. Our claims are therefore restricted to this audited family of sparse hexagonal graphs.

\section*{Data and Code Availability}
The benchmark generator, heuristic implementations, exact DFS oracle, and
evaluation scripts are available at \url{https://github.com/carsepmo/cpp_benchmarks/}. The version associated with this submission corresponds to GitHub release \texttt{v1.0.0}. The audited benchmark dataset is archived at Zenodo~\cite{sepulveda2026dataset}. The companion image archive used to generate the manuscript figures is archived at Zenodo~\cite{sepulveda2026images}.

\section{Conclusion}\label{sec:conclusion}

We presented a reproducible benchmark for coverage path planning on irregular
hexagonal grids motivated by maritime coverage scenarios. The benchmark combines
10\,000 audited instances, 17 deterministic heuristics, and a common evaluation
protocol spanning feasibility, path quality, and runtime. The main empirical
message is that relaxed complete coverage and Hamiltonian coverage are
qualitatively different tasks: methods that reliably solve the former generally
fail on the latter. Exact DFS confirms that the benchmark instances themselves
are Hamiltonian-feasible, so observed failures reflect algorithmic weakness
rather than hidden infeasibility. Among the tested classical baselines, the
strongest Hamiltonian performance is achieved by Warnsdorff-TI (index), while
the dominant implementation choice within the Warnsdorff family is the
residual-degree policy under endpoint reservation rather than tie-breaking
alone. We hope this benchmark provides a stronger baseline for future
heuristic, exact, and learning-based work on maritime coverage planning.

\balance

\bibliographystyle{IEEEtran}
\bibliography{reference_art}

\end{document}